\documentclass[12pt]{spieman}  

\usepackage{amsmath,amsfonts,amssymb}
\usepackage{graphicx}
\usepackage{setspace}
\usepackage{tocloft}

\usepackage{graphicx}
\usepackage{bm}
\usepackage{mathrsfs}
\usepackage{amsmath}
\usepackage{amssymb}
\usepackage{algorithm}
\usepackage{algorithmic}
\usepackage{array}
\usepackage{bm}
\usepackage{booktabs}
\usepackage{caption3}
\usepackage{url}
\usepackage{multirow}
\usepackage{xcolor}
\usepackage{lineno}
\newcommand{\etal}{\textit{et al.~}}
\newcommand{\ie}{i.e.}

\title{Learning efficient structured dictionary for image classification}

\author[a,b]{Zi-Qi Li}
\author[a,b,*]{Jun Sun}
\author[a,b]{Xiao-Jun Wu}
\author[a,b]{He-Feng Yin}
\affil[a]{Jiangnan University, School of Internet of Things Engineering, Lihu Avenue, Wuxi, China}
\affil[b]{Jiangsu Provincial Laboratory of Pattern Recognition and Computational Intelligence, Lihu Avenue, Wuxi, China}

\cftpagenumbersoff{figure}
\cftpagenumbersoff{table}

\begin{document}


\maketitle

\begin{abstract}
Recent years have witnessed the success of dictionary learning (DL) based approaches in the domain of pattern classification. In this paper, we present an efficient structured dictionary learning (ESDL) method which takes both the diversity and label information of training samples into account. Specifically, ESDL introduces alternative training samples into the process of dictionary learning. To increase the discriminative capability of representation coefficients for classification, an ideal regularization term is incorporated into the objective function of ESDL. Moreover, in contrast with conventional DL approaches which impose computationally expensive $\ell_1$-norm constraint on the coefficient matrix, ESDL employs $\ell_2$-norm regularization term. Experimental results on benchmark databases (including four face databases and one scene dataset) demonstrate that ESDL outperforms previous DL approaches. More importantly, ESDL can be applied in a wide range of pattern classification tasks.
\end{abstract}

\keywords{pattern classification; dictionary learning; ideal regularization term; label information}

{\noindent \footnotesize\textbf{*}Jun Sun,  \linkable{junsun@jiangnan.edu.cn} }

\begin{spacing}{2}   

\section{Introduction}
\label{sec-1}  
Dictionary learning (DL) has aroused considerable interest in recent years, and it has been successfully applied in various tasks, such as face recognition~\cite{chen2019noise}, image fusion~\cite{li2017multi} and person re-identification~\cite{zhu2017semi}. The most famous {\color{red}DL} method is the K-SVD algorithm~\cite{aharon2006k} which has been widely used in image compression and denoising. Nevertheless, K-SVD mainly focuses on the representational power of the dictionary without considering its capability for classification. To tackle this problem, Zhang \etal\cite{zhang2010discriminative} presented a discriminative K-SVD (D-KSVD) method by introducing the classification error into the framework of K-SVD. Jiang \etal\cite{jiang2011learning} further incorporated a label consistency constraint into K-SVD and proposed a label consistent K-SVD (LC-KSVD) algorithm. Kviatkovsky \etal\cite{kviatkovsky2016equivalence} mathematically proved the equivalence of the LC-KSVD and the D-KSVD algorithms up to a proper choice of regularization parameters. Zheng \etal\cite{zheng2015discriminative} developed a Fisher discriminative K-SVD (FD-KSVD) method which imposes Fisher discrimination criterion on the sparse coding coefficients. Similarly, by restricting the within-class scatter of a dictionary's representation coefficients, Xu \etal\cite{xu2016supervised} explored a supervised within-class-similar discriminative DL (SCDDL) algorithm. Motivated by the fact that kernel trick can capture the nonlinear similarity of features, Song \etal\cite{song2018euler} proposed an Euler label consistent K-SVD (ELC-KSVD) approach for image classification. By jointly learning a multi-class support vector machine (SVM) classifier, Cai \etal\cite{cai2014support} presented a support vector guided dictionary learning (SVGDL) model. To fully exploit the locality information of atoms in the learned dictionary, Yin \etal\cite{yin2019locality} proposed a locality constraint dictionary learning with support vector discriminative term (LCDL-SV) algorithm for pattern classification. {\color{red}Zhao \etal~\cite{zhao2015efficient} developed a DL algorithm which takes both sparsity and locality into consideration, and it selects part of the dictionary columns that are close to the input sample for coding.} Readers can refer to Ref. \citenum{xu2017survey} for a survey of {\color{red}DL} approaches.

However, conventional dictionary learning approaches do not fully exploit the diversity of training samples, especially when there are insufficient training samples. Moreover, the $\ell_0$ or $\ell_1$-norm constraint is often introduced to promote the sparsity of representation matrix, which is computationally expensive. To alleviate the above two problems, Xu \etal \cite{xu2017sample} proposed a new DL method in which the alternative training samples are introduced. The alternative training samples can be derived through the following two schemes, when providing insufficient training samples, virtual training samples can be generated and used as the alternative training samples. When we have large-scale training data, the whole training data can be divided into two parts with the same size, then one part is utilized as original training samples and the other part is the alternative training samples. {\color{red}Liu \etal~\cite{liu2020discriminative} developed a discriminative dictionary learning (DDL) algorithm based on sample diversity and locality of atoms.} Nevertheless, the label information of training samples is not exploited in Ref.~\citenum{xu2017sample} {\color{red}and DDL}, which undermines {\color{red}their} classification performance. To incorporate the label information of training samples into the formulation of DL, we introduce an ideal regularization term into the objective function of our proposed method. Through this term, representations of the training samples belonging to the same class are encouraged to be similar, which is beneficial for the subsequent classification stage. {\color{red}The demo code of our proposed ESDL is available at \url{https://github.com/li-zi-qi/ESDL}.}

Our main contributions can be summarized as follows.
\begin{itemize}
\item We take both the diversity and label information of training samples into account, and the introduced ideal regularization term associates the label information of training samples with that of atoms in the dictionary.
\item In a departure from conventional DL approaches which impose $\ell_1$-norm on the coefficient matrix, ESDL employs the $\ell_2$-norm constraint which is computationally efficient.
\item Our proposed ESDL is a general framework which can be applied in a wide range of pattern classification tasks.
\end{itemize}

The remainder of this paper is arranged as follows: Section \ref{sec_2} reviews the related work. Section \ref{sec_3} presents our proposed dictionary learning approach. Experimental results and analysis are presented in Section \ref{sec_4}. Finally, Section \ref{sec_5} concludes this paper.

\section{Related work}
\label{sec_2}
In this section, we briefly review the basic K-SVD \cite{aharon2006k} and its two discriminative extensions, \ie, D-KSVD \cite{zhang2010discriminative} and LC-KSVD~\cite{jiang2011learning}. Additionally, the dictionary learning method proposed by Xu \etal \cite{xu2017sample} is also introduced. We first give an introduction to the notations used throughout this paper. Let $\mathbf{Y}=\left[\boldsymbol{y}_{1}, \boldsymbol{y}_{2}, \ldots, \boldsymbol{y}_{N}\right] \in \mathbb{R}^{n \times N}$ be the data matrix of $N$ training samples belonging to $C$ classes, where $n$ is the dimension of vectorized data and $N$ is the total number of training samples, $\mathbf{D}=\left[\boldsymbol{d}_{1}, \boldsymbol{d}_{2}, \ldots, \boldsymbol{d}_{K}\right] \in \mathbb{R}^{n \times K}$ is the learned dictionary which has $K$ atoms, $\mathbf{X}=\left[\boldsymbol{x}_{1}, \boldsymbol{x}_{2}, \ldots, \boldsymbol{x}_{N}\right] \in \mathbb{R}^{K \times N}$ is the coding coefficients matrix of $\mathbf{Y}$ on the dictionary $\mathbf{D}$.

\subsection{K-SVD and its extensions}
\label{sec-3}
By generalizing the K-means clustering process, Aharon \etal \cite{aharon2006k} proposed K-SVD to learn an overcomplete dictionary that best approximates the given data. The objective function of K-SVD is formulated as follows,
\begin{equation} \label{eq-1}
    \min _{\mathbf{D}, \mathbf{X}}\|\mathbf{Y}-\mathbf{D} \mathbf{X}\|_{F}^{2}, \text { s.t. }\left\|\boldsymbol{x}_{i}\right\|_{0} \leq T_{0}
\end{equation}
where $\mathbf{D}$ is the dictionary that is to be learned, $\mathbf{X}$ is the coding coefficient matrix, and $T_{0}$ is a given sparsity level. Equation (\ref{eq-1}) can be solved by alternatively updating $\mathbf{D}$ and $\mathbf{X}$.

Although K-SVD achieves superb results in image denoising and restoration, its performance for classification is limited. To adapt K-SVD to classification tasks, Zhang \etal. \cite{zhang2010discriminative} developed D-KSVD algorithm by introducing the classification error term into the framework of K-SVD,
\begin{equation} \label{eq-2}
\min _{\mathbf{D}, \mathbf{W}, \mathbf{X}}\|\mathbf{Y}-\mathbf{D} \mathbf{X}\|_{F}^{2}+\beta\|\mathbf{H}-\mathbf{W} \mathbf{X}\|_{F}^{2}+\lambda\|\mathbf{W}\|_{F}^{2}, \text { s.t. }\left\|\boldsymbol{x}_{i}\right\|_{0} \leq T_{0}
\end{equation}
where $\mathbf{H}=\left[\boldsymbol{h}_{1}, \boldsymbol{h}_{2}, \ldots, \boldsymbol{h}_{N}\right] \in \mathbb{R}^{C \times N}$ is the label matrix of training data, $\boldsymbol{h}_{i}=[0,0, \ldots, 1, \ldots, 0,0]^{T} \in \mathbb{R}^{C \times 1}$ is the label vector of $\boldsymbol{y}_{i}$, and $\boldsymbol{W}$ is the parameters for a linear classifier. As can be seen from Eq. (\ref{eq-2}), dictionary and a linear classifier are jointly learned in D-KSVD. To further promote the discriminative capability of K-SVD, Jiang \etal \cite{jiang2011learning} presented LC-KSVD by solving the following optimization problem,
\begin{equation} \label{eq-3}
    \min _{\mathbf{D}, \mathbf{W}, \mathbf{A}, \mathbf{X}}\|\mathbf{Y}-\mathbf{D} \mathbf{X}\|_{F}^{2}+\alpha\|\mathbf{Q}-\mathbf{A} \mathbf{X}\|_{F}^{2}+\beta\|\mathbf{H}-\mathbf{W} \mathbf{X}\|_{F}^{2}, \text { s.t. }\left\|\boldsymbol{x}_{i}\right\|_{0} \leq T_{0}
\end{equation}
where $\mathbf{Q}=\left[\boldsymbol{q}_{1}, \boldsymbol{q}_{2}, \ldots, \boldsymbol{q}_{N}\right] \in \mathbb{R}^{K \times N}$ is an ideal representation matrix and $\mathbf{A}$ is a linear transformation matrix.

\subsection{Dictionary learning method proposed by Xu et al.}
\label{sec-4}
\begin{figure}[htbp]
	\centering
	\includegraphics[trim={0mm 0mm 0mm 0mm},clip, width = .5\textwidth]{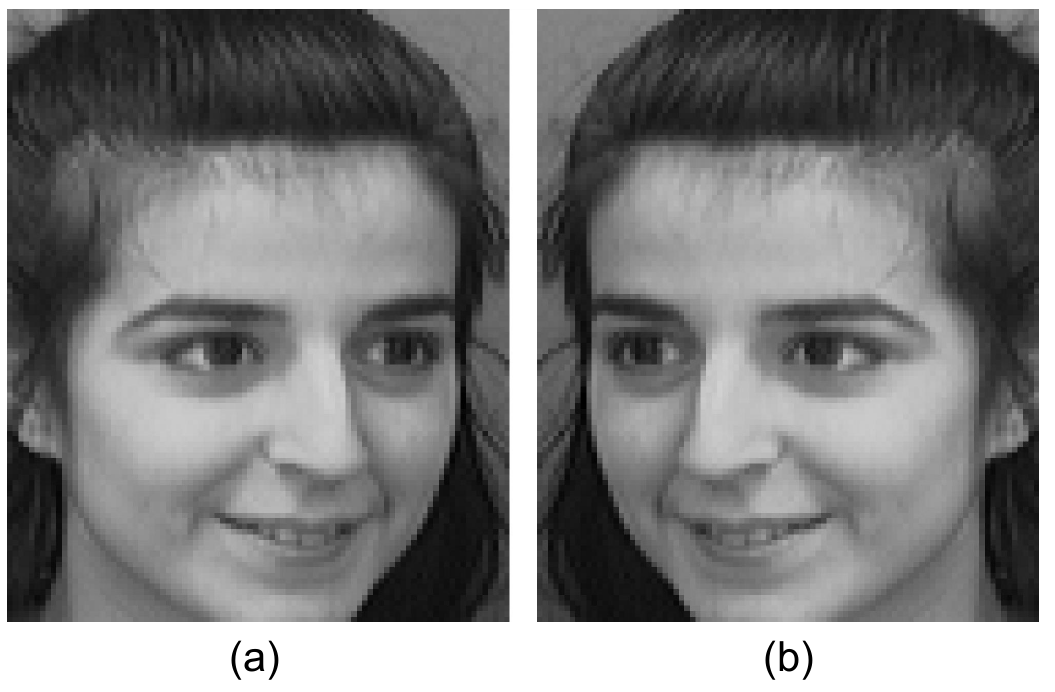}
	\caption{An illustration of alternative training sample. (a) One original face image; (b) Mirror face image of (a).}
	\label{fig:mirror}
\end{figure}

In order to promote the robustness of the learned dictionary to variations in the original training samples, such as illumination and expression changes in face recognition, Xu \etal \cite{xu2017sample} proposed a dictionary learning framework which takes the diversity of training samples into account, and the objective function is formulated as follows,
\begin{equation} \label{eq-4}
    \min _{\mathbf{D}, \mathbf{X}}\|\mathbf{Y}-\mathbf{D} \mathbf{X}\|_{F}^{2}+\alpha\left\|\mathbf{Y}_{alter}-\mathbf{D} \mathbf{X}\right\|_{F}^{2}+\beta\|\mathbf{X}\|_{F}^{2}, \text { s.t. }\left\|\boldsymbol{d}_{i}\right\|^{2}=1, i=1,2, \ldots, K
\end{equation}
where $\mathbf{Y}_{alter}$ is the data matrix for the alternative training data. For the scenario of insufficient training data, $\mathbf{Y}_{alter}$ can be obtained by generating virtual samples of the training samples. For instance, we can employ the mirror face images of the training data to form $\mathbf{Y}_{alter}$, and Fig.~\ref{fig:mirror} presents an original face image and its mirror face image, these two images belong to the same individual but they have different poses. Therefore, by introducing the mirror face images, diversity of training samples can be promoted to some extent. For large-scale training data, we can simply divide it into two parts with the same size and treat the first part and the second part as the original and virtual training samples, respectively.

\section{Proposed approach}
\label{sec_3}
\begin{figure}[htbp]
	\centering
	\includegraphics[trim={0mm 0mm 0mm 0mm},clip, width = .8\textwidth]{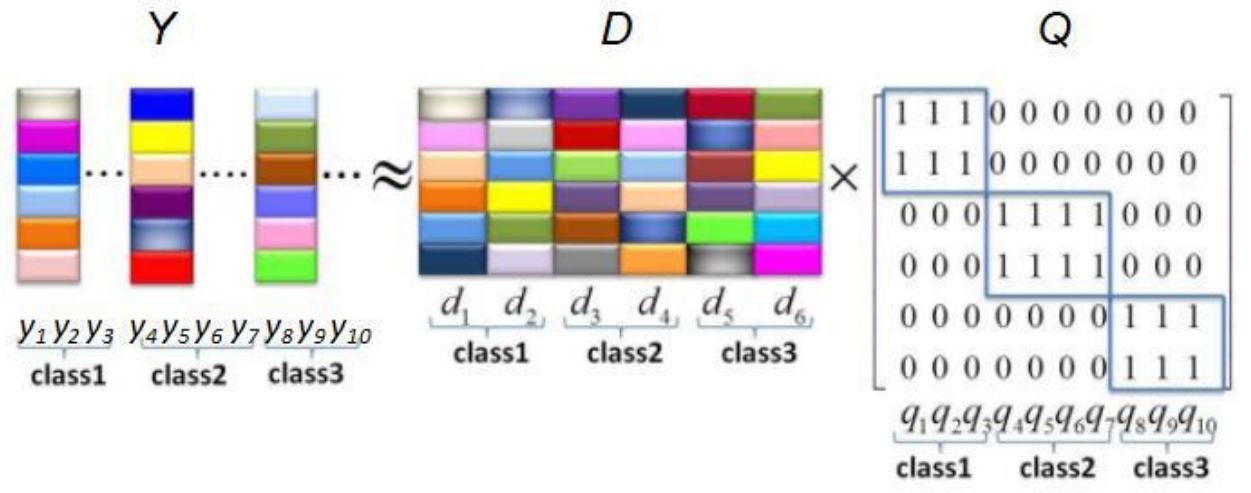}
	\caption{The ideal representation matrix $\mathbf{Q}$ for the training data $\mathbf{Y}$.}
	\label{fig:illus_Q}
\end{figure}

From the formulation of Eq. (\ref{eq-4}), we can see that {\color{red}the label information of training data is not exploited} in the process of dictionary learning, which leads to limited performance for pattern classification. For classification tasks, utilization of label information of training data can bring improved results. {\color{red}Furthermore, with respect to a semantic dictionary $\mathbf{D}$, the optimal coefficient matrix $\mathbf{X}$ for $\mathbf{Y}$ should be block-diagonal~\cite{liu2012robust}:
\begin{equation*} 
\mathbf{X}^*=\begin{bmatrix}
 \mathbf{X}_1^*& \boldsymbol{0} & \boldsymbol{0} & \boldsymbol{0}\\ 
 \boldsymbol{0}& \mathbf{X}_2^* & \boldsymbol{0} & \boldsymbol{0}\\ 
\boldsymbol{0} & \boldsymbol{0} & \cdots  & \boldsymbol{0}\\ 
\boldsymbol{0} & \boldsymbol{0} & \boldsymbol{0} & \mathbf{X}_C^*
\end{bmatrix}
\end{equation*}
During the update process, the quality of the dictionary $\mathbf{D}$ is affected by the discrimination of the coefficient matrix $\mathbf{X}$ . Consequently, to enhance the discriminative capability of coefficient matrix during the dictionary learning process, we employ the class labels of training data to design a discriminative term. As in Ref.~\citenum{zhang2013learning}, we introduce matrix $\mathbf{Q}$ in block-diagonal form as an ideal representation. }
{\color{red}Let} $\mathbf{Q}=\left[\boldsymbol{q}_{1}, \boldsymbol{q}_{2}, \ldots, \boldsymbol{q}_{N}\right] \in \mathbb{R}^{K \times N}$ {\color{red}be} an ideal representation matrix formed by the label information of training data and dictionary atoms, $\boldsymbol{q}_{i}=[0,0, \ldots, 1,1, \ldots, 0,0]^{T} \in \mathbb{R}^{K \times 1}$. The entries in $\boldsymbol{q}_{i}$ are 1 when the training samples and the dictionary atoms have the same class label. An illustration of $\mathbf{Q}$ is shown in Fig.~\ref{fig:illus_Q}, suppose $\mathbf{Y}=\left[\boldsymbol{y}_{1}, \boldsymbol{y}_{2}, \ldots, \boldsymbol{y}_{10}\right]$ and $\mathbf{D}=\left[\boldsymbol{d}_{1}, \boldsymbol{d}_{2}, \ldots, \boldsymbol{d}_{6}\right]$, where $\boldsymbol{y}_{1}$, $\boldsymbol{y}_{2}$ and $\boldsymbol{y}_{3}$ belong to the first class, $\boldsymbol{y}_{4}$, $\boldsymbol{y}_{5}$, $\boldsymbol{y}_{6}$ and $\boldsymbol{y}_{7}$ belong to the second class, and $\boldsymbol{y}_{8}$, $\boldsymbol{y}_{9}$ and $\boldsymbol{y}_{10}$ belong to the third class. $\mathbf{D}$ has 3 sub-dictionaries and each has 2 atoms. As we can see from Fig.~\ref{fig:illus_Q}, $\mathbf{Q}$ exhibits a block-diagonal structure.

{\color{red}To obtain a coefficient matrix $\mathbf{X}$ close to $\mathbf{Q}$,we add a regularization term $\left \| \mathbf{X}-\mathbf{Q} \right \|_F^2$ to include the structured and discriminative information from training samples into the dictionary learning process.} The objective function of our proposed ESDL is formulated as,
\begin{equation} \label{eq:obj_pro}
	\min _{\mathbf{D}, \mathbf{X}}\|\mathbf{Y}-\mathbf{D} \mathbf{X}\|_{F}^{2}+\alpha\left\|\mathbf{Y}_{alter}-\mathbf{D} \mathbf{X}\right\|_{F}^{2}+\beta\|\mathbf{X}\|_{F}^{2}+\gamma\|\mathbf{X}-\mathbf{Q}\|_{F}^{2}, \text { s.t. }\left\|d_{i}\right\|^{2}=1, i=1, \ldots, K
\end{equation}
{\color{red}where $\alpha$, $\beta$, and $\gamma$ are balancing parameters, and the regularization term $\left \| \mathbf{X}-\mathbf{Q} \right \|_F^2$ encourages samples from the same class to have similar representations, which results in good classification performance even using a simple linear classifier.}

\subsection{Optimization}
\label{sec-6}
We employ alternative strategy to optimize Eq. (\ref{eq:obj_pro}), \ie, update one variable when the other is fixed. The detailed updating procedures are presented as follows.

Update $\mathbf{X}$: when $\mathbf{D}$ is fixed, Eq. (\ref{eq:obj_pro}) degenerates into the following problem,
\begin{equation} \label{eq:update_X}
	\min _{\mathbf{X}}\|\mathbf{Y}-\mathbf{D} \mathbf{X}\|_{F}^{2}+\alpha\left\|\mathbf{Y}_{alter}-\mathbf{D} \mathbf{X}\right\|_{F}^{2}+\beta\|\mathbf{X}\|_{F}^{2}+\gamma\|\mathbf{X}-\mathbf{Q}\|_{F}^{2}
\end{equation}

Eq. (\ref{eq:update_X}) has the following closed-form solution,
\begin{equation} \label{eq:solu_X}
	\mathbf{X}=\left(\mathbf{D}^{T} \mathbf{D}+\alpha \mathbf{D}^{T} \mathbf{D}+\beta \mathbf{I}+\gamma \mathbf{I}\right)^{-1}\left(\mathbf{D}^{T} \mathbf{Y}+\alpha \mathbf{D}^{T} \mathbf{Y}_{alter}+\gamma \mathbf{Q}\right)
\end{equation}

Update $\mathbf{D}$:  when $\mathbf{X}$ is fixed, $\mathbf{D}$ can be updated by solving the following problem,
\begin{equation} \label{eq:update_D}
\min _{\mathbf{D}}\|\mathbf{Y}-\mathbf{D} \mathbf{X}\|_{F}^{2}+\alpha\left\|\mathbf{Y}_{alter}-\mathbf{D} \mathbf{X}\right\|_{F}^{2}
\end{equation}

Eq. (\ref{eq:update_D}) has the following closed-form solution,
\begin{equation} \label{eq:solu_D}
\mathbf{D}=\left(\mathbf{Y} \mathbf{X}^{T}+\alpha \mathbf{Y}_{alter} \mathbf{X}^{T}\right)\left(\mathbf{X} \mathbf{X}^{T}+\alpha \mathbf{X} \mathbf{X}^{T}\right)^{-1}
\end{equation}
At the beginning of optimization of Eq. (\ref{eq:obj_pro}), dictionary $\mathbf{D}$ is initialized via K-SVD, \ie, K-SVD is performed on the training data of each class to obtain a sub-dictionary, then all the sub-dictionaries are concatenated to form the whole dictionary. Based on the label information of training samples and atoms in the dictionary, the ideal representation matrix $\mathbf{Q}$ can be constructed. Then Eq. (\ref{eq:obj_pro}) can be optimized by iteratively updating $\mathbf{X}$ and $\mathbf{D}$. The complete optimization process of Eq.~(\ref{eq:obj_pro}) is outlined in Algorithm~\ref{alg1}.

\begin{algorithm} 
\caption{Optimization process of Eq.~(\ref{eq:obj_pro})} 
\label{alg1} 
\begin{algorithmic}[1]
\REQUIRE Training data matrix $\mathbf{Y}$, alternative training data $\mathbf{Y}_{alter}$, parameters $\alpha$, $\beta$, and $\gamma$.
\STATE Initialize $\mathbf{D}$ via K-SVD, construct the ideal representation matrix $\mathbf{Q}$;
\WHILE{not converged} 
\STATE Update $\mathbf{X}$ by Eq.~(\ref{eq:solu_X});
\STATE Update $\mathbf{D}$ by Eq.~(\ref{eq:solu_D});
\ENDWHILE 
\ENSURE The learned dictionary $\mathbf{D}$ and the coefficient matrix $\mathbf{X}$ of training data.
\end{algorithmic} 
\end{algorithm}

\subsection{Classification Scheme}
\label{sec-7}
When the dictionary learning process is completed, the learned dictionary $\mathbf{D}$ and representation matrix $\mathbf{X}$ of training data are obtained. Based on the representation matrix $\mathbf{X}$ and label matrix $\mathbf{H}$ of training data, a linear classifier can be learned by solving the following problem,
\begin{equation} \label{eq-10}
	\mathbf{W}=\arg \min _{\mathbf{W}}\|\mathbf{H}-\mathbf{W} \mathbf{X}\|_{F}^{2}+\lambda\|\mathbf{W}\|_{F}^{2}
\end{equation}

Eq. (\ref{eq-10}) has closed-form solution, which is formulated as,
\begin{equation} \label{eq:solu_class}
	\mathbf{W}=\mathbf{H X}^{T}\left(\mathbf{X} \mathbf{X}^{T}+\lambda \mathbf{I}\right)^{-1}
\end{equation}

{\color{red}As in Ref.~\citenum{xu2017sample}}, to classify a test sample $\boldsymbol{y}$, first we obtain its coefficient vector $\boldsymbol{x}$ via orthogonal matching pursuit (OMP) algorithm~\cite{tropp2007signal}, then the label for $\boldsymbol{y}$ is given by,
\begin{equation} \label{eq-13}
\textrm{identity}(\boldsymbol{y})=\arg \max _{k}\left(\boldsymbol{g}_{k}\right), \textrm{where} \  \boldsymbol{g}=\mathbf{W} \boldsymbol{x}
\end{equation}

The classification procedures of our proposed method are summarized in Algorithm~\ref{alg2}.

\begin{algorithm} 
\caption{Classification process of our proposed method} 
\label{alg2} 
\begin{algorithmic}[1]
\REQUIRE The learned dictionary $\mathbf{D}$, the coefficient matrix $\mathbf{X}$ of training data, label matrix $\mathbf{H}$ of training data and test sample $\boldsymbol{y}$.
\STATE Obtain the linear classifier $\mathbf{W}$ via~(\ref{eq:solu_class});
\STATE Compute the coding vector $\boldsymbol{x}$ of test sample $\boldsymbol{y}$ via OMP;
\STATE Calculate $\boldsymbol{g}=\mathbf{W} \boldsymbol{x}$;
\ENSURE $\textrm{identity}(\boldsymbol{y})=\arg \max _{k}\left(\boldsymbol{g}_{k}\right)$.
\end{algorithmic} 
\end{algorithm}

\subsection{{\color{red}SDL with $\ell_1$-norm constraint}}
{\color{red}
In our proposed ESDL, the $\ell_2$-norm constraint is imposed on the coefficient matrix to facilitate the process of dictionary learning. For completeness, we present SDL-$\ell_1$ which imposes the $\ell_1$-norm constraint on the coefficient matrix. The objective function of SDL-$\ell_1$ is formulated as follows,
\begin{equation} 
\label{eq:obj_L1}
\min _{\mathbf{D}, \mathbf{X}}\|\mathbf{Y}-\mathbf{D} \mathbf{X}\|_{F}^{2}+\alpha\left\|\mathbf{Y}_{alter}-\mathbf{D} \mathbf{X}\right\|_{F}^{2}+\beta\|\mathbf{X}\|_{1}+\gamma\|\mathbf{X}-\mathbf{Q}\|_{F}^{2}, \text { s.t. }\left\|d_{i}\right\|^{2}=1, i=1, \ldots, K
\end{equation}
where $\|\mathbf{X}\|_{1}$ is the $\ell_1$-norm of $\mathbf{X}$.

To solve the above problem, by introducing an auxiliary variable $\mathbf{Z}$, Eq.~(\ref{eq:obj_L1}) can be converted into the following equivalent optimization problem,
\begin{equation} 
\label{eq:obj_equi}
\begin{split}
&\min _{\mathbf{D}, \mathbf{X},\mathbf{Z}}\|\mathbf{Y}-\mathbf{D} \mathbf{X}\|_{F}^{2}+\alpha\left\|\mathbf{Y}_{alter}-\mathbf{D} \mathbf{X}\right\|_{F}^{2}+\beta\|\mathbf{Z}\|_{1}+\gamma\|\mathbf{X}-\mathbf{Q}\|_{F}^{2}, \\
&\text { s.t. }\mathbf{X}=\mathbf{Z}, \ \left\|d_{i}\right\|^{2}=1
\end{split}
\end{equation}
The alternating direction method of multipliers (ADMM) scheme~\cite{boyd2011distributed} can be adopted to solve Eq.~(\ref{eq:obj_equi}), and the augmented Langrangian function is formulated as,

\begin{equation} 
\label{eq:langr}
\begin{split}
&\mathcal{L}(\mathbf{D}, \mathbf{X},\mathbf{Z},\mathbf{L},\mu)=\|\mathbf{Y}-\mathbf{D} \mathbf{X}\|_{F}^{2}+\alpha\left\|\mathbf{Y}_{alter}-\mathbf{D} \mathbf{X}\right\|_{F}^{2}+\beta\|\mathbf{Z}\|_{1}+\\
&\gamma\|\mathbf{X}-\mathbf{Q}\|_{F}^{2}+<\mathbf{L},\mathbf{X}-\mathbf{Z}>+\frac{\mu}{2}\left \| \mathbf{X}-\mathbf{Z} \right \|_F^2
\end{split}
\end{equation}
where $<\mathbf{A},\mathbf{B}>=\textrm{trace}(\mathbf{A}^T\mathbf{B})$, $\mathbf{L}$ is the Lagrange multiplier, and $\mu>0$ is a penalty parameter. Eq.~(\ref{eq:langr}) can be solved iteratively by updating $\mathbf{X}$, $\mathbf{Z}$ and $\mathbf{D}$ once at a time. The detailed procedures are presented as follows.

Update $\mathbf{X}$: Fix the other variables and update $\mathbf{X}$ by solving the following problem,
\begin{equation} 
\label{eq:pro_X}
\min _{\mathbf{X}}\|\mathbf{Y}-\mathbf{D} \mathbf{X}\|_{F}^{2}+\alpha\left\|\mathbf{Y}_{alter}-\mathbf{D} \mathbf{X}\right\|_{F}^{2}+\gamma\|\mathbf{X}-\mathbf{Q}\|_{F}^{2}+\frac{\mu}{2}\left \| \mathbf{X}-\mathbf{Z}+\frac{\mathbf{L}}{\mu} \right \|_F^2
\end{equation}
which has the following closed-form solution,
\begin{equation} 
\label{eq:solu_X1}
\mathbf{X}=\left(\mathbf{D}^{T} \mathbf{D}+\alpha \mathbf{D}^{T} \mathbf{D}+\frac{\mu}{2} \mathbf{I}+\gamma \mathbf{I}\right)^{-1}\left(\mathbf{D}^{T} \mathbf{Y}+\alpha \mathbf{D}^{T} \mathbf{Y}_{alter}+\gamma \mathbf{Q}+\frac{\mu\mathbf{Z}-\mathbf{L}}{2}\right)
\end{equation}

Update $\mathbf{Z}$: When the other variables are fixed, Eq.~(\ref{eq:langr}) with respect to $\mathbf{Z}$ is boiled down to the following problem,
\begin{equation} 
\label{eq:pro_Z}
\min _{\mathbf{Z}}\beta\|\mathbf{Z}\|_{1}+\frac{\mu}{2}\left \| \mathbf{Z}-(\mathbf{X}+\frac{\mathbf{L}}{\mu}) \right \|_F^2
\end{equation}
The closed-form solution of Z is given by,
\begin{equation} 
\label{eq:solu_Z}
\mathbf{Z}=\textrm{max}(\mathbf{X}+\frac{\mathbf{L}}{\mu}-\frac{\beta}{\mu},0)+\textrm{min}(\mathbf{X}+\frac{\mathbf{L}}{\mu}+\frac{\beta}{\mu},0)
\end{equation}

Update $\mathbf{D}$: The optimization problem of dictionary $\mathbf{D}$ has the same formulation as Eq.~(\ref{eq:update_D}) , and its closed-form solution is shown in Eq.~(\ref{eq:solu_D}).

The complete procedures of solving Eq.~(\ref{eq:langr}) are summarized in Algorithm~\ref{alg3}. From Algorithm~\ref{alg3}, we can see that there are five variables to be updated in the optimization process of SDL-$\ell_1$, while there are only two variables to be updated in our proposed ESDL. Therefore, ESDL is more computationally efficient than SDL-$\ell_1$ in the training phase, which is also demonstrated by the following experiments.}

\begin{algorithm} 
\caption{Optimization process of Eq.~(\ref{eq:langr})} 
\label{alg3} 
\begin{algorithmic}[1]
\REQUIRE Training data matrix $\mathbf{Y}$, alternative training data $\mathbf{Y}_{alter}$, parameters $\alpha$, $\beta$, and $\gamma$.
\STATE Initialize $\mathbf{D}$ via K-SVD, $\mathbf{Z}=\mathbf{L}=\mathbf{0}$, and construct the ideal representation matrix $\mathbf{Q}$, $\mu=0.01$, $\mu_{\textrm{max}}=10^8$, $\rho=1.1$;
\WHILE{not converged} 
\STATE Update $\mathbf{X}$ by Eq.~(\ref{eq:solu_X1});
\STATE Update $\mathbf{Z}$ by Eq.~(\ref{eq:solu_Z});
\STATE Update $\mathbf{D}$ by Eq.~(\ref{eq:solu_D});
\STATE Update $\mathbf{L}$ by $\mathbf{L}=\mathbf{L}+\mu(\mathbf{X}-\mathbf{Z})$;
\STATE Update $\mu$ by $\mu=\textrm{min}(\rho \mu,\mu_{\textrm{max}})$;
\STATE  Check the convergence condiction.
\ENDWHILE 
\ENSURE The learned dictionary $\mathbf{D}$ and the coefficient matrix $\mathbf{X}$ of training data.
\end{algorithmic} 
\end{algorithm}

\section{Experimental results and analysis}
\label{sec_4}
In this section, we evaluate the classification performance of our proposed ESDL on five benchmark datasets: the Extended Yale B database, the AR database, the PIE database, the LFW database, and the Scene 15 dataset. To illustrate the superiority of ESDL, we compare ESDL with the following approaches: SRC~\cite{wright2008robust}, CRC~\cite{zhang2011sparse}, K-SVD~\cite{aharon2006k}, D-KSVD~\cite{zhang2010discriminative}, LC-KSVD~\cite{jiang2011learning}, SVDGL~\cite{cai2014support}, the method in Ref.~\citenum{xu2017sample} {\color{red}and SDL-$\ell_1$}. There are three parameters in our proposed ESDL, {\color{red}\ie, $\alpha$, $\beta$ and $\gamma$, and they are tuned to achieve the best performance via fivefold cross-validation from [$10^{-4}$, $10^{-3}$, 0.01, 0.1].} On the four face databases, $\alpha$, $\beta$ and $\gamma$ are set to {\color{red}0.01}, $10^{-3}$ and $10^{-3}$, respectively, while on the Scene 15 dataset, $\alpha$, $\beta$ and $\gamma$ are set to 0.1, $10^{-4}$ and $10^{-4}$, respectively. Apart from the recognition accuracy, we also present the training time and testing time (in seconds) of all the competing methods. All experiments are run with MATLAB R2019b under Windows 10 on a PC equipped with Intel i9-8950HK 2.90 GHz CPU and 32 GB RAM.

\subsection{Experiments on the Extended Yale B Database}
\label{sec-9}
\begin{figure}[htbp]
	\centering
	\includegraphics[trim={0mm 0mm 0mm 0mm},clip, width = .8\textwidth]{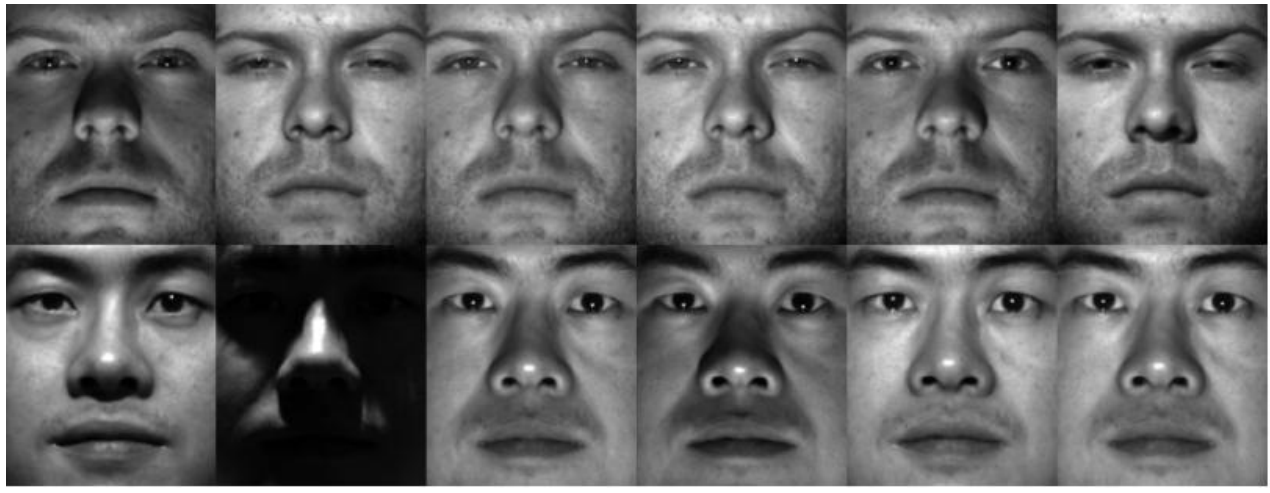}
	\caption{Example images from the Extended Yale B database.}
	\label{fig:exam_eyaleb}
\end{figure}

There are 2414 face images of 38 subjects in the Extended Yale B database, and these images have variations in illumination, some example images are shown in Fig.~\ref{fig:exam_eyaleb}. Each individual contains 59-64 images. In our experiments, all images are resized to 32$\times$32 pixels. Twenty images per person (the first five images per person are always selected, and the other fifteen images per person are randomly selected from the remaining of the images) are used as training samples and the rest as test samples. {\color{red}The dictionary size $K$ is set to 760 in this experiment.} We repeat the experiments ten times and record the average recognition accuracy. Experimental results are summarized in Table~\ref{tab:tab-1}. It can be seen that the proposed algorithm achieves a higher average recognition accuracy than its competing approaches. Moreover, ESDL is very efficient in terms of training and testing time. {\color{red}Specifically, ESDL is 11 times faster than SDL-$\ell_1$ in the training stage.}

\begin{figure}[htbp]
	\centering
	\includegraphics[trim={0mm 0mm 0mm 0mm},clip, width = .8\textwidth]{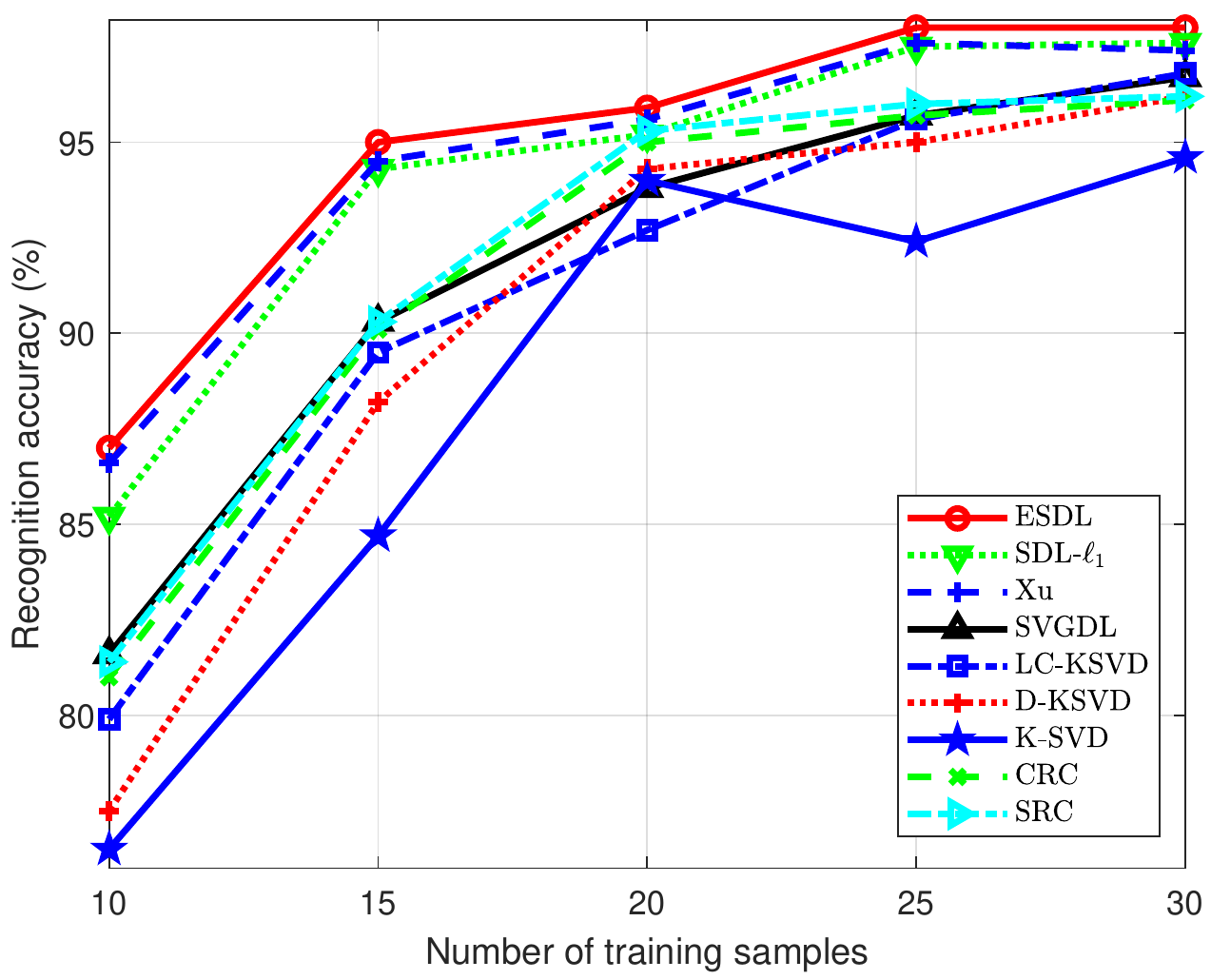}
	\caption{{\color{red}Recognition accuracy versus number of training samples per subject on the Extended Yale B database.}}
	\label{fig:vary_train}
\end{figure}

{\color{red}To examine how the number of training samples influences the recognition accuracy of compared approaches, we randomly select 10, 15, 20, 25 and 30 images per subject for training, and the remaining for testing. The averaged recognition accuracy of 10 runs are plotted against the number of training samples per subject in Fig.~\ref{fig:vary_train}. It can be seen that ESDL consistently outperforms the others in all cases.}

\begin{table}[] \label{tab:tab-1}
	\centering
	\caption{Recognition results on the Extended Yale B database.}
	\begin{tabular}{cccc}
		\hline
		Methods & Accuracy (\%)& Training time (s) & Testing time (s) \\
		\hline
		SRC~\cite{wright2008robust}     & 95.3     &     No Need    &      1.56           \\
		CRC~\cite{zhang2011sparse}     & 95.0     &      No Need  &       0.82          \\
		K-SVD~\cite{aharon2006k}   & 94.0     &      1.82            &      0.35           \\
		D-KSVD~\cite{zhang2010discriminative}  & 94.3     &     22.80             &     0.42            \\
		LC-KSVD~\cite{jiang2011learning} & 92.7     &      38.08            &      0.43           \\
		SVGDL~\cite{cai2014support}   & 93.8     &  43.82  &    0.12         \\
		Xu's DL~\cite{xu2017sample} & 95.6     &     2.87             &       0.56          \\
{\color{red}SDL-$\ell_1$}    &  {\color{red}95.6}    &  {\color{red}30.87}   &  {\color{red}0.47}	\\
		ESDL    & \textbf{95.9}     &      2.72            &        0.44        	\\
		\hline
	\end{tabular}
\end{table}

\subsection{Experiments on the AR Database}
\label{sec-10}
\begin{figure}[htbp]
	\centering
	\includegraphics[trim={0mm 0mm 0mm 0mm},clip, width = .8\textwidth]{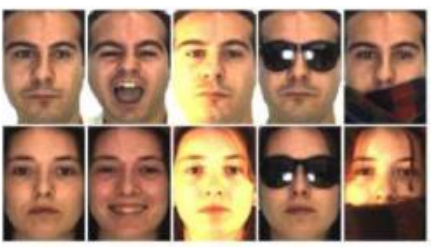}
	\caption{Example images from the AR database.}
	\label{fig:exam_ar}
\end{figure}

The AR face database includes over 4000 images of 126 subjects, each subject has 26 images collected in two separate sessions which vary in expression, illumination and disguise (wearing sunglasses of scarves), example images are shown in Fig.~\ref{fig:exam_ar}. As the experimental setting in {\color{red}Ref.~\citenum{xu2017sample}}, we use a subset of the AR face database containing 3120 images for 120 subjects (65 men and 55 women). The size of the images is 40$\times$50 pixels.

In our experiments, seven undisguised images in session 1 and one occluded image per subject are used as training samples (the first face image per subject with sunglasses in sessions 1 and 2 are not used as training samples). Therefore, sixteen images per subject (seven undisguised images in session 2 and the rest nine occluded images) are used for testing. {\color{red}In the experiment, the number of dictionary atoms $K$ is 960.} Experiments are repeated for ten times and the average recognition results are shown in Table~\ref{tab-2}. One can see that ESDL outperforms the others in accuracy, and its training and testing time are comparable to those of Xu's method~\cite{xu2017sample}. ESDL is 437 times faster than SVGDL for the training phase, about 24 times faster than LC-KSVD.

\begin{table}[] \label{tab-2}
	\centering
	\caption{Recognition results on the AR database.}
	\begin{tabular}{cccc}
		\hline
		Methods & Accuracy (\%)& Training time (s) & Testing time (s) \\
		\hline
		SRC~\cite{wright2008robust}     & 72.2     &     No Need             &     65.19            \\
		CRC~\cite{zhang2011sparse}     & 71.4     &     No Need              &    53.36             \\
		K-SVD~\cite{aharon2006k}   & 78.8     &    3.40              &     0.44            \\
		D-KSVD~\cite{zhang2010discriminative}  & 74.1     &     69.03             &     0.70            \\
		LC-KSVD~\cite{jiang2011learning} & 74.2     &     131.76             &    0.64             \\
		SVGDL~\cite{cai2014support}   & 78.0     &     2403.67             &    0.22             \\
		Xu's DL~\cite{xu2017sample} & 79.8     &    5.74              &     0.77            \\
{\color{red}SDL-$\ell_1$}    & {\color{red}80.2}  &  {\color{red}61.06} & {\color{red}0.69} 	\\
		ESDL    & \textbf{80.2}     &   5.50               &     0.64           	\\
		\hline
	\end{tabular}
\end{table}

\subsection{Experiments on the PIE Database}
\label{sec-11}
\begin{figure}[htbp]
	\centering
	\includegraphics[trim={0mm 0mm 0mm 0mm},clip, width = .8\textwidth]{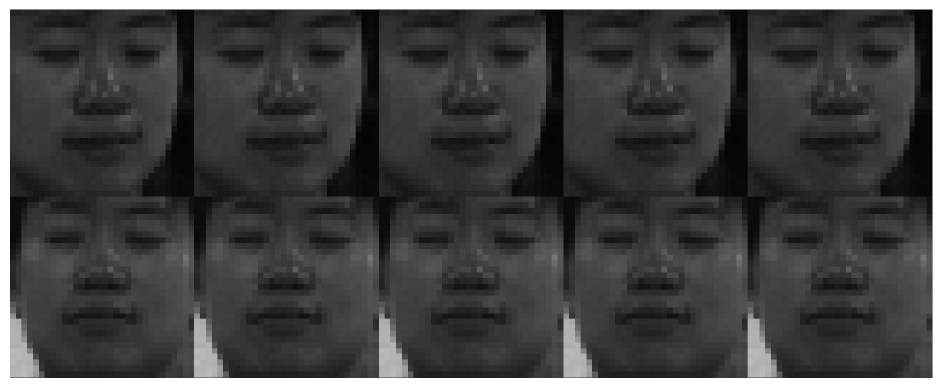}
	\caption{Example images from the PIE database.}
	\label{fig:exam_pie}
\end{figure}

The PIE database contains 41,368 front-face images of 68 subjects, and the images of each subject are captured under 13 different poses, 43 different illumination conditions, and 4 different facial expressions, example images from this database are depicted in Fig.~\ref{fig:exam_pie}.

Following the common experimental settings, we choose the five near-frontal poses (C05, C07, C09, C27, C29) of each subject and use all the images under different illumination conditions and facial expressions. Thus we obtain 170 images for each subject. Each image is normalized to the size of 32$\times$32 pixels. Ten images per subject (including the first five images) are randomly selected as training samples and the remaining as test samples. {\color{red}The dictionary size $K$ is set to 680.} Experiments are repeated for ten times the the average results are listed in Table~\ref{tab-3}. It can be observed that ESDL achieves the highest accuracy. Specifically, it outperforms Xu's~\cite{xu2017sample} method and SVGDL by 0.6\% and 1.2\%, respectively. Meanwhile, ESDL is 426 times faster than SVGDL.

\begin{table}[] \label{tab-3}
	\centering
	\caption{Recognition results on the PIE database.}
	\begin{tabular}{cccc}
		\hline
		Methods & Accuracy (\%)& Training time (s) & Testing time (s) \\
		\hline
		SRC~\cite{wright2008robust}     & 72.1     &     No Need              &  56.49               \\
		CRC~\cite{zhang2011sparse}     & 74.1     &     No Need              &  51.04               \\
		K-SVD~\cite{aharon2006k}   & 72.0     &      1.76            &    1.50             \\
		D-KSVD~\cite{zhang2010discriminative}  & 71.9     &     19.05             &   2.56              \\
		LC-KSVD~\cite{jiang2011learning} & 72.3     &     30.33             &     2.63            \\
		SVGDL~\cite{cai2014support}   & 76.4     &   986.03               &   0.17              \\
		Xu's DL~\cite{xu2017sample} & 77.0     &    2.32              &   2.58              \\
{\color{red}SDL-$\ell_1$}    & {\color{red}76.6}     & {\color{red}23.09} & {\color{red}2.82} \\
		ESDL    & \textbf{77.6}     &     2.31             &        2.61        	\\
		\hline
	\end{tabular}
\end{table}

\subsection{Experiments on the LFW Database}
\label{sec_4-4}
\begin{figure}[htbp]
	\centering
	\includegraphics[trim={0mm 0mm 0mm 0mm},clip, width = .8\textwidth]{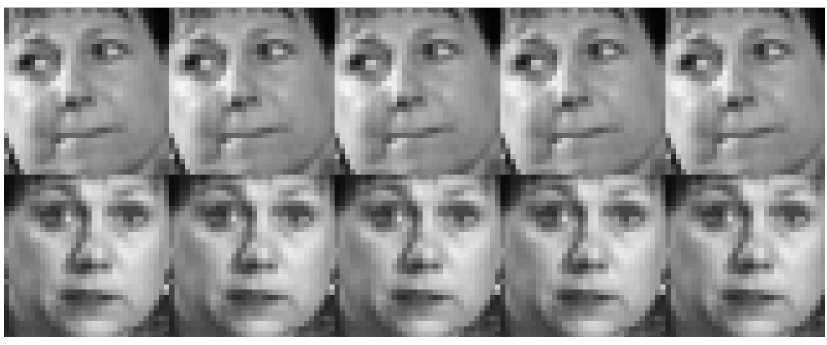}
	\caption{Example images from the LFW database.}
	\label{fig:exam_lfw}
\end{figure}

We use a subset of the LFW database which contains 1215 images of 86 individuals, {\color{red}each image is resized to 32$\times$32 pixels.} Example images from this database are shown in Fig.~\ref{fig:exam_lfw}. In our experiments, six images per person are randomly selected as training samples and the remaining as test samples. {\color{red}The number of dictionary atoms $K$ is set to 430 in the experiment.} Experiments are repeated for ten times and the average recognition accuracy is summarized in Table~\ref{tab-4}. It can be seen from Table~\ref{tab-4} that the performance gains of our proposed ESDL is significant on this dataset. It outperforms Xu's method by 2.3\%, and it is very efficient in terms of training time.

\begin{table}[] \label{tab-4}
	\centering
	\caption{Recognition results on the LFW database.}
	\begin{tabular}{cccc}
		\hline
		Methods & Accuracy (\%)& Training time (s) & Testing time (s) \\
		\hline
		SRC~\cite{wright2008robust}     &  30.3    &    No Need               &     3.28            \\
		CRC~\cite{zhang2011sparse}     & 24.8     &    No Need               &      0.49           \\
		K-SVD~\cite{aharon2006k}   &  29.7    &     1.65             &   0.10              \\
		D-KSVD~\cite{zhang2010discriminative}  &  19.6    &     8.51             &     0.14            \\
		LC-KSVD~\cite{jiang2011learning} &  19.4    &      11.03            &    0.14             \\
		SVGDL~\cite{cai2014support}   &   29.1   &      606.48            &        0.02         \\
		Xu's DL~\cite{xu2017sample} & 33.1     &      0.93            &    0.14             \\
{\color{red}SDL-$\ell_1$}    & {\color{red}34.4}     & {\color{red}7.75} & {\color{red}0.17} \\
		ESDL    & \textbf{35.4}     &    0.93              &   0.13             	\\
		\hline
	\end{tabular}
\end{table}

\subsection{Experiments on the Scene 15 Dataset}
\label{sec-13}
\begin{figure}[htbp]
	\centering
	\includegraphics[trim={0mm 0mm 0mm 0mm},clip, width = .8\textwidth]{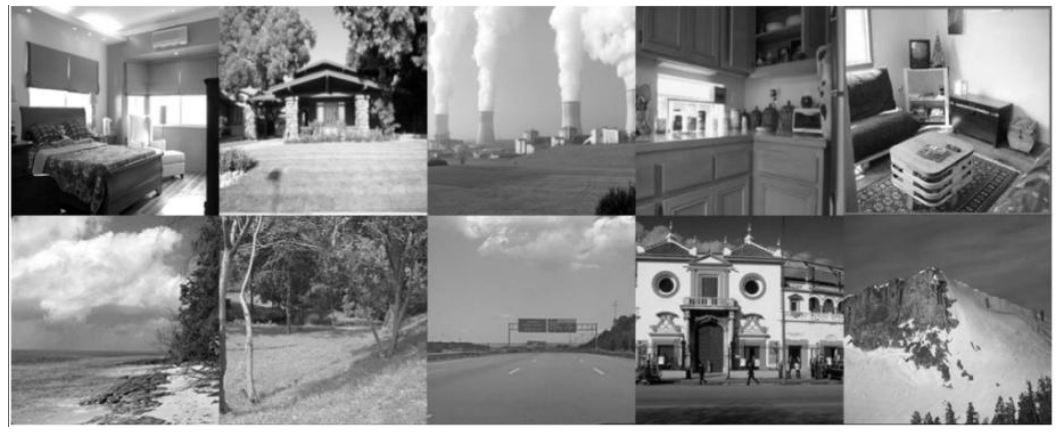}
	\caption{Example images from the Scene 15 dataset.}
	\label{fig:exam_scene}
\end{figure}

\begin{figure}[htbp]
	\centering
	\includegraphics[trim={0mm 0mm 0mm 0mm},clip, width = .8\textwidth]{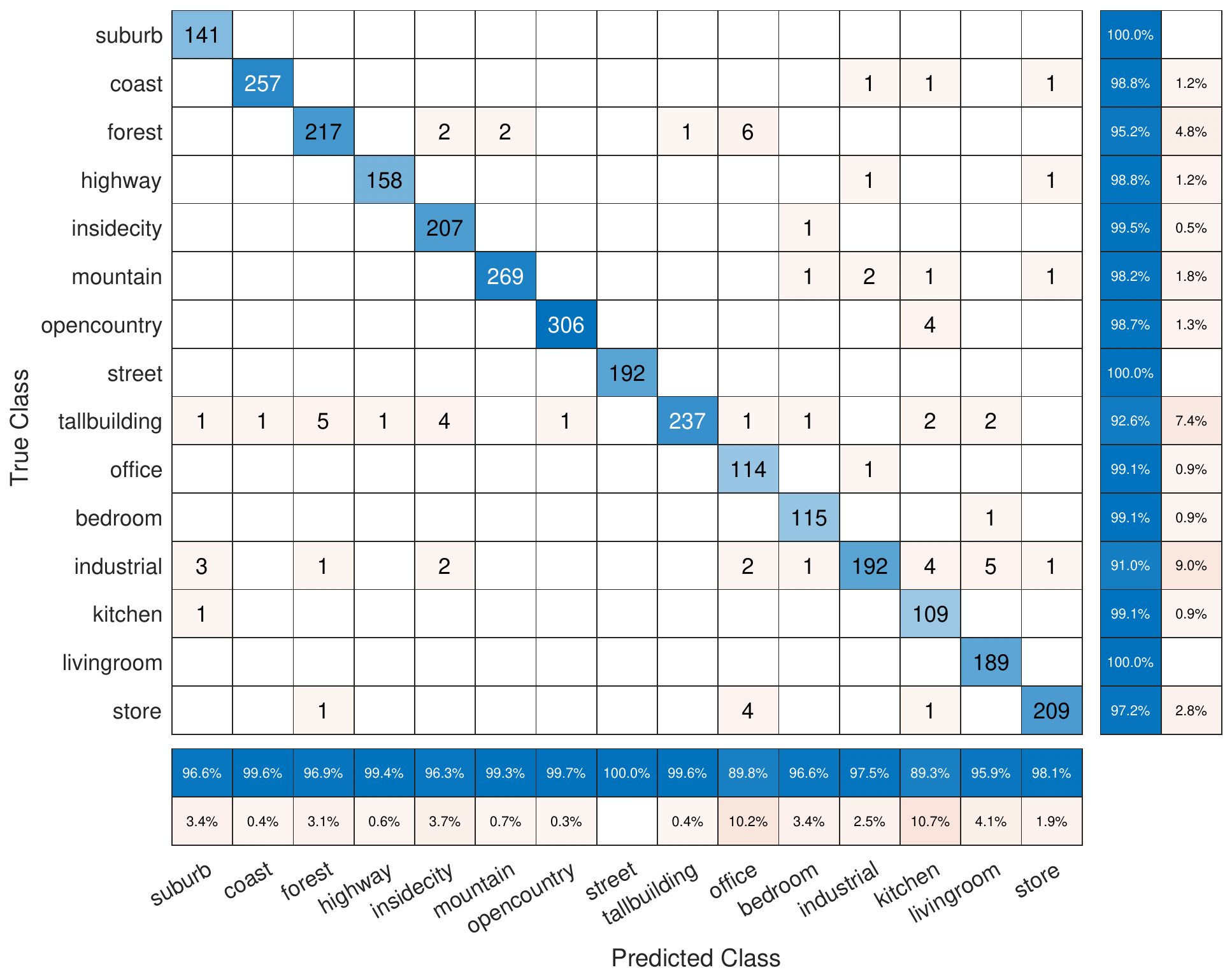}
	\caption{Confusion matrix on the Scene 15 dataset.}
	\label{fig:confusion}
\end{figure}

Scene 15 dataset has 15 natural scene categories, which comprises a wide range of indoor and outdoor scenes, such as bedroom, office and mountain, example images from this dataset are shown in Fig.~\ref{fig:exam_scene}. Each category has 200-400 images, and the average image size is about 250$\times$300 pixels. For fair comparison, we employ the 3000-dimensional SIFT-based features used in LC-KSVD~\cite{jiang2011learning}. According to the experimental settings in Ref.~\citenum{xu2017sample}, 100 images per category are randomly selected as training data and the rest as test data. For the training samples, the first 50 images per category are used as original training samples and the other 50 images as alternative training samples. The number of atoms in the learned dictionary is 450. Experimental results are listed in Table~\ref{tab-5}. We can observe that our proposed ESDL is superior to other approaches, and it is about 130 times faster than SVGDL. We also plot the confusion matrix for ESDL in Fig.~\ref{fig:confusion}, in which diagonal elements are well-marked. It can be seen that ESDL obtains 100\% recognition accuracy for categories of suburb, street, and livingroom. Experiments on the Scene 15 dataset demonstrate that ESDL is not only suitable for face recognition, but for scene categorization as well. Actually, our proposed ESDL is a general framework, which can be applied to other pattern classification tasks.

\begin{table}[] \label{tab-5}
	\centering
	\caption{Recognition results on the Scene 15 dataset.}
	\begin{tabular}{cccc}
		\hline
		Methods & Accuracy (\%)& Training time (s) & Testing time (s) \\
		\hline
		SRC~\cite{wright2008robust}     & 91.8     &     No Need              &     52.68             \\
		CRC~\cite{zhang2011sparse}     & 95.8     &    No Need               &     45.26           \\
		K-SVD~\cite{aharon2006k}   & 86.7     &      9.05            &     0.49            \\
		D-KSVD~\cite{zhang2010discriminative}  & 89.1     &      63.44            &    0.52             \\
		LC-KSVD~\cite{jiang2011learning} & 92.9     &     77.86             &   0.53              \\
		SVGDL~\cite{cai2014support}   & 95.4     &     226.07             &   0.08              \\
		Xu's DL~\cite{xu2017sample} & 97.5     &      1.86            &    0.56             \\
{\color{red}SDL-$\ell_1$}    & {\color{red}97.4}  &  {\color{red}17.73}  & {\color{red}0.55}   \\
		ESDL    & \textbf{97.7}     &   1.74     &     0.57   \\
		\hline
	\end{tabular}
\end{table}

\subsection{Parameter Analysis}

\begin{figure}[htbp]
	\centering
	\includegraphics[trim={0mm 0mm 0mm 0mm},clip, width = .8\textwidth]{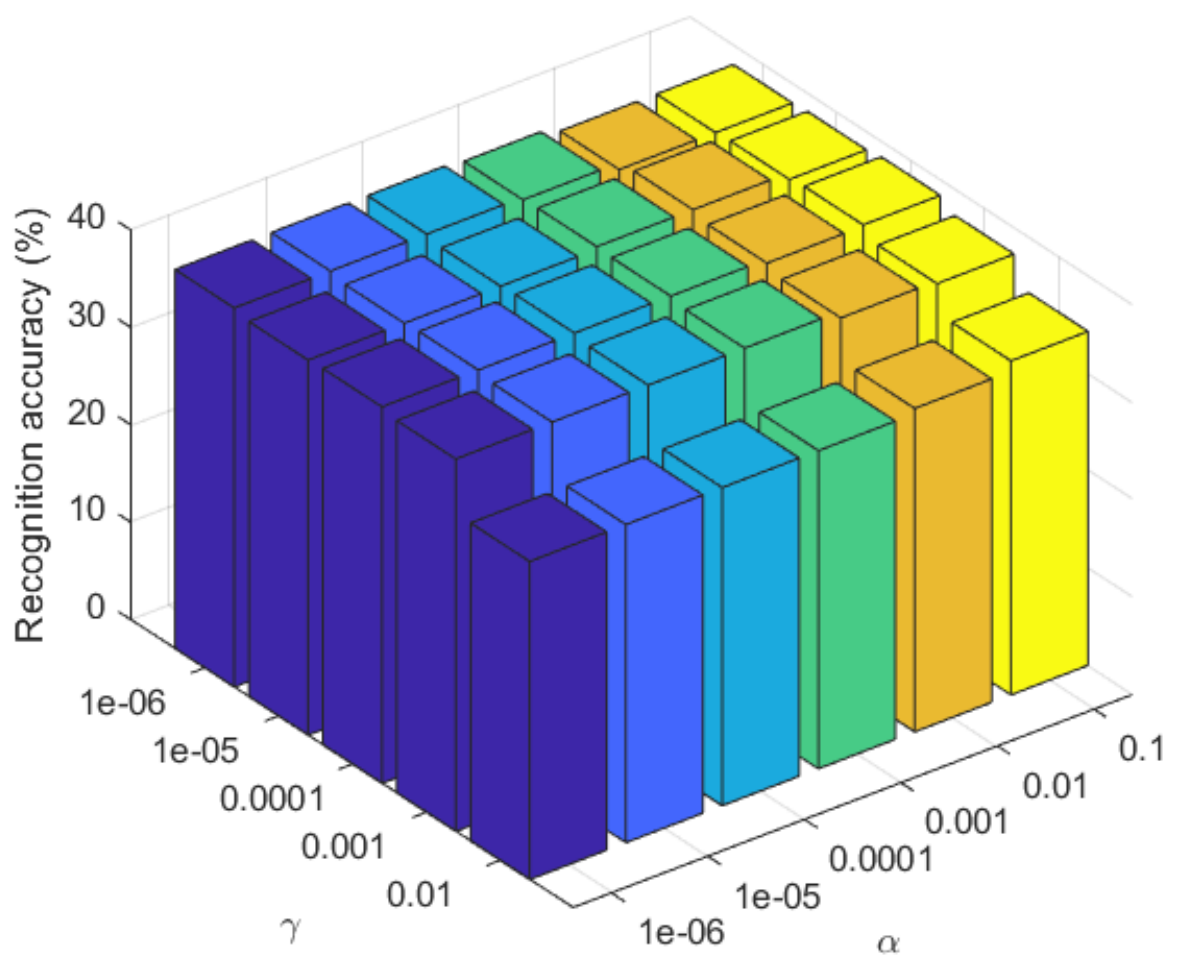}
	\caption{Recognition accuracy of ESDL versus parameters $\alpha$ and $\gamma$ on the LFW database.}
	\label{fig:parameter}
\end{figure}

There are three parameters in the formulation of our proposed ESDL, \ie, $\alpha$, $\beta$ and $\gamma$ in Eq.~(\ref{eq:obj_pro}), $\beta$ is usually set to a relatively small value (1e-4 or 1e-3 ) in our experiments
. To examine how the remaining parameters $\alpha$ and $\gamma$ influence the performance of ESDL, we conduct experiments on the LFW database. Experimental setting is the same as in Section~\ref{sec_4-4} and the number of training samples per subject is 6. Fig.~\ref{fig:parameter} illustrates the effect of parameter selection. One can see that the recognition performance of ESDL is stable when the value of parameter $\alpha$ varies in quite a wide range, \ie, $[10^{-6},0.1]$. Meanwhile, ESDL achieves better accuracy when $\gamma$ has relatively small value, \ie, $[10^{-6},10^{-3}]$. According to the above experimental results, we set $\alpha={\color{red}0.01}$ and $\gamma=0.001$ on the LFW database.

\section{Conclusions}
\label{sec_5}
We proposed an efficient structured dictionary learning (ESDL) method in which both the diversity and label information of training samples are considered. By introducing the ideal regularization term, label information of training data and the dictionary atoms are associated. Moreover, ESDL imposes $\ell_2$-norm constraint instead of the $\ell_1$-norm on the coefficient matrix which makes ESDL computationally efficient. It is worth noting that our proposed ESDL can be applied in a variety of pattern classification tasks. By generating virtual training samples, ESDL can tackle the problem of insufficient training samples. When providing large-scale training samples, we can simply divide the original training set into two parts and feed them into the framework of ESDL. Experimental results on five well known datasets demonstrate the superiority of ESDL over some state-of-the-art DL approaches.

In this paper, we did not explicitly consider the situation that both the training and test samples are contaminated due to occlusion or corruption, thus in future, we will introduce low rank matrix recovery (LRMR) technique into ESDL to tackle the above scenarios.

\acknowledgments
This work was supported in part by the National Natural Science Foundation of China (Projects Numbers: 61673194, 61672263, 61672265, U1836218 and 61876072), and in part by the national first-class discipline program of Light Industry Technology and Engineering (Project Number: LITE2018-25).


\bibliography{report}   

\begin{thebibliography}{10}

\bibitem{chen2019noise}
Z.~Chen, X.-J. Wu, H.-F. Yin, {\em et~al.}, ``Noise-robust dictionary learning
  with slack block-diagonal structure for face recognition,'' {\em Pattern
  Recognition} , 107118  (2019).

\bibitem{li2017multi}
H.~Li and X.-J. Wu, ``Multi-focus image fusion using dictionary learning and
  low-rank representation,'' in {\em International Conference on Image and
  Graphics},  675--686, Springer  (2017).

\bibitem{zhu2017semi}
X.~Zhu, X.-Y. Jing, L.~Yang, {\em et~al.}, ``Semi-supervised cross-view
  projection-based dictionary learning for video-based person
  re-identification,'' {\em IEEE Transactions on Circuits and Systems for Video
  Technology} {\bf 28}(10), 2599--2611  (2017).

\bibitem{aharon2006k}
M.~Aharon, M.~Elad, and A.~Bruckstein, ``K-svd: An algorithm for designing
  overcomplete dictionaries for sparse representation,'' {\em IEEE Transactions
  on signal processing} {\bf 54}(11), 4311--4322  (2006).

\bibitem{zhang2010discriminative}
Q.~Zhang and B.~Li, ``Discriminative k-svd for dictionary learning in face
  recognition,'' in {\em 2010 IEEE Computer Society Conference on Computer
  Vision and Pattern Recognition},  2691--2698, IEEE  (2010).

\bibitem{jiang2011learning}
Z.~Jiang, Z.~Lin, and L.~S. Davis, ``Learning a discriminative dictionary for
  sparse coding via label consistent k-svd,'' in {\em CVPR 2011},  1697--1704,
  IEEE  (2011).

\bibitem{kviatkovsky2016equivalence}
I.~Kviatkovsky, M.~Gabel, E.~Rivlin, {\em et~al.}, ``On the equivalence of the
  lc-ksvd and the d-ksvd algorithms,'' {\em IEEE transactions on pattern
  analysis and machine intelligence} {\bf 39}(2), 411--416  (2016).

\bibitem{zheng2015discriminative}
H.~Zheng and D.~Tao, ``Discriminative dictionary learning via fisher
  discrimination k-svd algorithm,'' {\em Neurocomputing} {\bf 162}, 9--15
  (2015).

\bibitem{xu2016supervised}
L.~Xu, X.~Wu, K.~Chen, {\em et~al.}, ``Supervised within-class-similar
  discriminative dictionary learning for face recognition,'' {\em Journal of
  Visual Communication and Image Representation} {\bf 38}, 561--572  (2016).

\bibitem{song2018euler}
Y.~Song, Y.~Liu, Q.~Gao, {\em et~al.}, ``Euler label consistent k-svd for image
  classification and action recognition,'' {\em Neurocomputing} {\bf 310},
  277--286  (2018).

\bibitem{cai2014support}
S.~Cai, W.~Zuo, L.~Zhang, {\em et~al.}, ``Support vector guided dictionary
  learning,'' in {\em European Conference on Computer Vision},  624--639,
  Springer  (2014).

\bibitem{yin2019locality}
H.-F. Yin, X.-J. Wu, and S.-G. Chen, ``Locality constraint dictionary learning
  with support vector for pattern classification,'' {\em IEEE Access} {\bf 7},
  175071--175082  (2019).

\bibitem{zhao2015efficient}
Z.~Zhao and G.~Feng, ``Efficient algorithm for sparse coding and dictionary
  learning with applications to face recognition,'' {\em Journal of Electronic
  Imaging} {\bf 24}(2), 023009--023009  (2015).

\bibitem{xu2017survey}
Y.~Xu, Z.~Li, J.~Yang, {\em et~al.}, ``A survey of dictionary learning
  algorithms for face recognition,'' {\em IEEE access} {\bf 5}, 8502--8514
  (2017).

\bibitem{xu2017sample}
Y.~Xu, Z.~Li, B.~Zhang, {\em et~al.}, ``Sample diversity, representation
  effectiveness and robust dictionary learning for face recognition,'' {\em
  Information Sciences} {\bf 375}, 171--182  (2017).

\bibitem{liu2020discriminative}
S.~Liu, Y.~Wang, X.~Wu, {\em et~al.}, ``Discriminative dictionary learning
  algorithm based on sample diversity and locality of atoms for face
  recognition,'' {\em Journal of Visual Communication and Image Representation}
  , 102763  (2020).

\bibitem{liu2012robust}
G.~Liu, Z.~Lin, S.~Yan, {\em et~al.}, ``Robust recovery of subspace structures
  by low-rank representation,'' {\em IEEE transactions on pattern analysis and
  machine intelligence} {\bf 35}(1), 171--184  (2012).

\bibitem{zhang2013learning}
Y.~Zhang, Z.~Jiang, and L.~S. Davis, ``Learning structured low-rank
  representations for image classification,'' in {\em Proceedings of the IEEE
  conference on computer vision and pattern recognition},  676--683  (2013).

\bibitem{tropp2007signal}
J.~A. Tropp and A.~C. Gilbert, ``Signal recovery from random measurements via
  orthogonal matching pursuit,'' {\em IEEE Transactions on information theory}
  {\bf 53}(12), 4655--4666  (2007).

\bibitem{boyd2011distributed}
S.~Boyd, N.~Parikh, E.~Chu, {\em et~al.}, ``Distributed optimization and
  statistical learning via the alternating direction method of multipliers,''
  {\em Foundations and Trends{\textregistered} in Machine learning} {\bf 3}(1),
  1--122  (2011).

\bibitem{wright2008robust}
J.~Wright, A.~Y. Yang, A.~Ganesh, {\em et~al.}, ``Robust face recognition via
  sparse representation,'' {\em IEEE transactions on pattern analysis and
  machine intelligence} {\bf 31}(2), 210--227  (2008).

\bibitem{zhang2011sparse}
L.~Zhang, M.~Yang, and X.~Feng, ``Sparse representation or collaborative
  representation: Which helps face recognition?,'' in {\em 2011 International
  conference on computer vision},  471--478, IEEE  (2011).

\end{thebibliography}
\bibliographystyle{spiejour}   


\vspace{2ex}\noindent\textbf{Zi-Qi Li} received his B.E. degree in School of Information Engineering, Yangzhou University, Yangzhou, China, in 2012. Currently, he is pursuing the Ph.D. degree in School of Internet of Things Engineering, Jiangnan University, Wuxi, China. His research interests include feature representation, information fusion, dictionary learning and low rank representation.

\vspace{1ex}\noindent\textbf{Jun Sun} received his PhD in control theory and engineering, and an MSc in Computer Science and Technology from Jiangnan University, China, in 2009 and 2003, respectively. He is currently working as a full Professor with the Department of Computer Science and Technology, Jiangnan University, China. He is also vice director of Jiangsu Provincial Engineering Laboratory of Pattern Recognition and Computational Intelligence, Jiangsu Province. His major research areas and work are related to computational intelligence, machine learning, bioinformatics, among others. He published more than 150 papers in journals, conference proceedings and several books in the above areas.

\vspace{1ex}\noindent\textbf{Xiao-Jun Wu} received the B.S. degree in mathematics from Nanjing Normal University, Nanjing, China, in 1991. He received the M.S. degree in 1996, and the Ph.D. degree in pattern recognition and intelligent systems in 2002, both from Nanjing University of Science and Technology, Nanjing, China. He joined Jiangnan University in 2006, where he is currently a Professor. He has published more than 200 papers in his fields of research. He was a visiting researcher in the Centre for Vision, Speech, and Signal Processing (CVSSP), University of Surrey, U.K., from 2003 to 2004. His current research interests include pattern recognition, computer vision, fuzzy systems, neural networks, and intelligent systems.

\vspace{1ex}\noindent\textbf{He-Feng Yin} received his B.S. degree in School of Computer Science and Technology from Xuchang University, Xuchang, China, in 2011. Currently, he is a PhD candidate in School of IoT Engineering, Jiangnan University, Wuxi, China. He was a visiting PhD student at centre for vision, speech and signal processing (CVSSP), University of Surrey, under the supervision of Prof. Josef Kittler. His research interests include representation based classification methods, dictionary learning and low rank representation.

\listoffigures
\listoftables

\end{spacing}
\end{document}